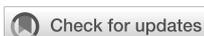

# Vision meets algae: A novel way for microalgae recognization and health monitor

Shizheng Zhou[1], Juntao Jiang[2], Xiaohan Hong[3], Pengcheng Fu[1] and Hong Yan[1]*

[1]State Key Laboratory of Marine Resource Utilization in South China Sea, Hainan University, Haikou, China, [2]College of Control Science and Engineering, Zhejiang University, Hangzhou, China, [3]College of Engineering and Applied Sciences, The State University of New York at Stony Brook, Stony Brook, NY, United States

Marine microalgae are widespread in the ocean and play a crucial role in the ecosystem. Automatic identification and location of marine microalgae in microscopy images would help establish marine ecological environment monitoring and water quality evaluation system. We proposed a new dataset for the detection of marine microalgae and a range of detection methods, the dataset including images of different genus of algae and the same genus in different states. We set the number of unbalanced classes in the data set and added images of mixed water samples in the test set to simulate the actual situation in the field. Then we trained, validated and tested the, TOOD, YOLOv5, YOLOv8 and variants of RCNN algorithms on this dataset. The results showed both one-stage and two-stage object detection models can achieve high mean average precision, which proves the ability of computer vision in multi-object detection of microalgae, and provides basic data and models for real-time detection of microalgal cells.

KEYWORDS

marine microalgae, object detection, data annotation, microscopic imaging, computer vision

# 1 Introduction

Marine microalgae are widely distributed in the ocean (Lu et al., 2021a) as part of the "blue carbon sink" that uses solar energy and dissolves $CO_2$ to produce oxygen as well as carbohydrates through photosynthesis (Lu et al., 2021b). They are thus involved in the global ocean–atmosphere carbon cycle to mitigate anthropogenic $CO_2$ emission, which is the leading cause of escalating climate change. Healthy and viable algal growth is critical to the prosperity of diverse marine ecosystems in the ocean and carbon capture, utilization, and storage. Also, species and quantities of microalgae are widely used as indicators for marine ecological environment monitoring and water quality evaluation worldwide (Gilbert et al., 1992; Domenighini and Giordano, 2009; Gordon and Leggat, 2010; Parmar et al., 2016; Naughton et al., 2020; Peter et al., 2021).

At present, the method of identifying and counting algal species is mainly through manual observation under a microscope, which is time-consuming and laborious and relies on their expertise, and algal samples are very likely to die in the process of collection and transportation, as shown in Figure 1A. The methods to monitor the health of algal cells at the single-cell level include chlorophyll determination, flow cytometry, and molecular biology. However, these methods rely on complex sensors and large instruments, which are not easily popularized in the general laboratory. It is challenging to conduct real-time detection when sampling in the field, as shown in Figure 1B.

In order to monitor and determine the health status of marine microalgae *in situ*, hardware and software integration is needed to enable real-time algal image acquisition, data analysis, and implementation of machine learning algorithms to recognize and classify the cells of marine microalgae. Microfluidic technology is a promising approach to recognizing microalgae and monitoring the health of microalgae at the single-cell level. It can achieve high throughput, has good biocompatibility, provides the ability to integrate with other methods, and requires a small sample size. Combining microfluidic technology with microscopic imaging can avoid the potential damage of labeled cells and enable dynamic cell detection, as shown in Figure 2. The method would produce images of algal cells or videos of algal flowing, just enough to provide data for computer vision methods.

Compared to current manual microscopic identification, which has disadvantages such as high professional level requirements, discontinuity of classifiers, and being time-consuming, automatic marine microalgae identification by using computer vision methods (Coltelli et al., 2014; Promdaen et al., 2014; Giraldo-Zuluaga et al., 2018; Reimann et al., 2020; Barsanti et al., 2021; Cao et al., 2021; Park et al., 2021; Xu et al., 2021) can meet the needs of rapid monitoring and provide convenience for researchers in marine and environmental science. In this analysis of microalgal images, automatic localization and identification are expected to be achieved simultaneously, facilitating the downstream cell analysis. As the main tasks of classification and localization, object detection can provide the basis for algae identification based on image information combined with biomorphological features.

Object detection is the task of precisely estimating the concepts and locations of objects in each image (Felzenszwalb et al., 2010; Zhao et al., 2019). Traditional object detectors (Viola and Jones, 2001a; Viola and Jones, 2001b; Dalal and Triggs, 2005; Felzenszwalb et al., 2008) based on sophisticated handcrafted features used to be mainstream methods before the popularity of deep learning. Thanks to the capability of learning robust and high-level feature representations of images, convolutional neural networks (CNNs) have been widely used in object detection. Two-stage detectors (Girshick, 2015; Girshick et al., 2015; He et al., 2015; Ren et al., 2017; Cai and Vasconcelos, 2018; Sun et al., 2020; Zhang et al., 2020a) achieved object region selection and detection in two steps while one-stage detectors (Liu et al., 2016; Redmon et al., 2016; Lin et al., 2017; Redmon and Farhadi, 2017; Redmon and Farhadi, 2018; Bochkovskiy et al., 2020; Law and Deng, 2020; Feng et al., 2021; Ge et al., 2021; Jocher, 2021) gave the class probabilities and position coordinate values of objects directly. Recently, some transformer-based methods (Carion et al., 2020) were proposed, showing their strong ability in object detection. Commonly used object datasets include Microsoft COCO (Lin et al., 2014) and Pascal VOC (Everingham et al., 2010).

Some existing related works on microalgal microscopy images applied such a deep learning-based method. Cao et al. (2021) used MobileNet (Howard et al., 2017) and SPP (He et al., 2015) to improve YOLOv3 while Park et al. (2021) trained the YOLOv3

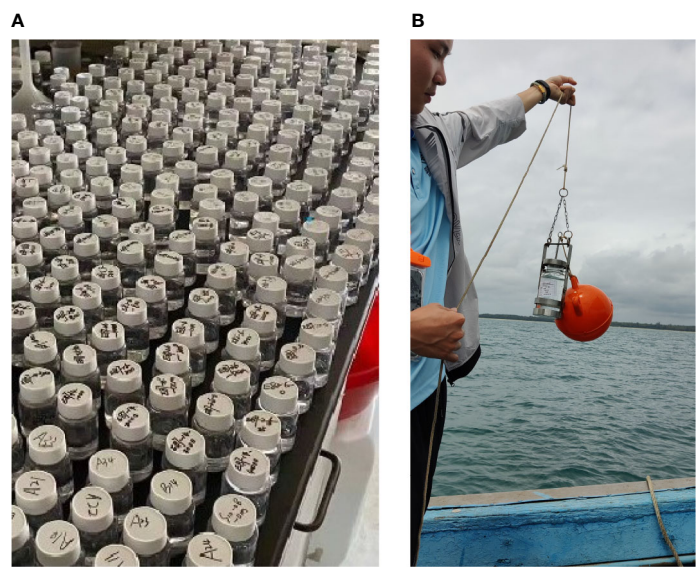

**A**   **B**

FIGURE 1
Detection and identification of microalgal cells. **(A)** Algal samples awaiting identification. Manual identification of algal water samples is time-consuming and laborious. **(B)** Algal water samples are collected in the field. The collected algal samples generally need to be fixed and concentrated, stored in cold storage, and then transported to scientific research institutions.

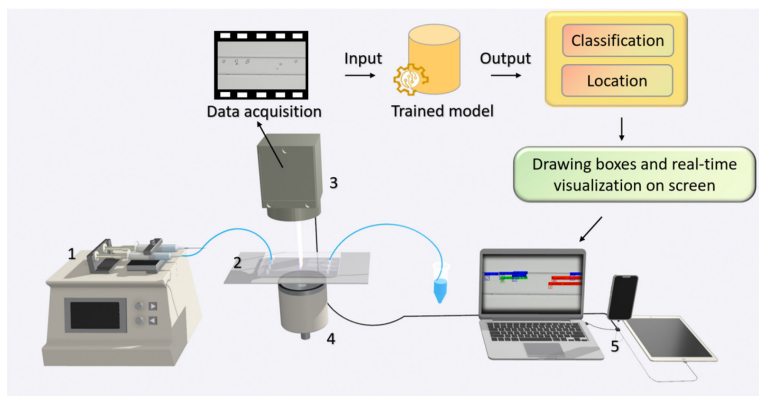

FIGURE 2
Microscopic imaging combined with computer vision and microfluidic technology is used for real-time detection and health monitor of algae. (1) Syringe pump; (2) microfluidic chip; (3) CMOS microscope; (4) light source; (5) computer.

model with Darknet53 backbone (Redmon and Farhadi, 2018). Cao acquired microalgal images using a Nikon Ts2-FL/TS2 fluorescence microscope and then used the data augmentation methods to obtain many images (Cao et al., 2021). The microalgal species in this dataset include *Prorocentrum lima* and *Karenia mikimotoi*. Park et al. (Redmon and Farhadi, 2018) collected 1,114 algal images with 3,663 objects collected by a microscope (Eclipse Ni, NIKON, Japan). These images were split into 10, 20, or 30 genera to compare the model's performance with different numbers of objects for classification.

In this study, we have taken samples of six genera of microalgae commonly found in the ocean (*Pinnularia*, *Chlorella*, *Platymonas*, *Dunaliella salina*, *Isochrysis*, and *Symbiodinium*) and created an image dataset for them. We also collected the images of *Symbiodinium* in different physiological states known as normal, bleaching, and translating, and an image set mixing all microalgal genera described above. We provided many details about this dataset, such as producing process, the original resolution, and the total number of images and objects. Since we will subsequently apply this dataset to the competition, the number of images containing *Pinnularia* was intentionally set lower than the others to examine how well the participants handled the sample imbalance problem in the dataset. In addition, we added an image set of mixed samples to the test set, which was designed better to fit the algal samples in the field environment.

We hope to discover minimal differences between cells from a computer vision perspective and understand life's heterogeneity, randomness, and synergy at the single-cell level. We gave a brief description of the dataset information and the annotation process.

Additionally, we trained some classical or state-of-the-art object detection methods as baselines on the training set and got a test result on the test set. Researchers are welcome to develop their own marine microalgal object detection algorithms on this dataset and compare their results with these baselines. We sincerely hope that this dataset of microscopy images for marine microalga detection will boost related research on marine biology and help establish future real-time monitoring and water quality evaluation system of the marine environment.

# 2 Materials and methods

This section specifically describes the workflow of building this dataset. The workflow of the dataset production includes data (water samples containing different genera of microalgae) collection, microscopic imaging, image annotation, annotation proofreading, and splitting of the dataset into the training set and the test set, as shown in Figure 3. In addition, we also described the training environment and relevant evaluation metrics of the object detection models.

## 2.1 Data collection

Microalgal samples collected in this study are common genera in seawater, including *Platymonas*, *Isochrysis*, *Chlorella*, *D. salina*, *Pinnularia*, and *Symbiodinium*, as shown in Figure 4. The microscopic magnification of these cell images is 40×. The algal

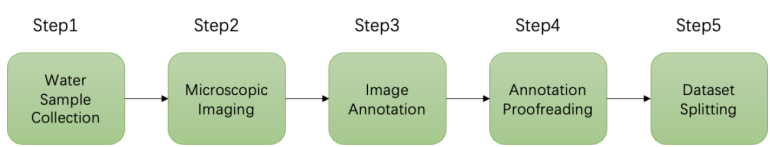

FIGURE 3
The workflow of producing the dataset for marine microalgae detection.

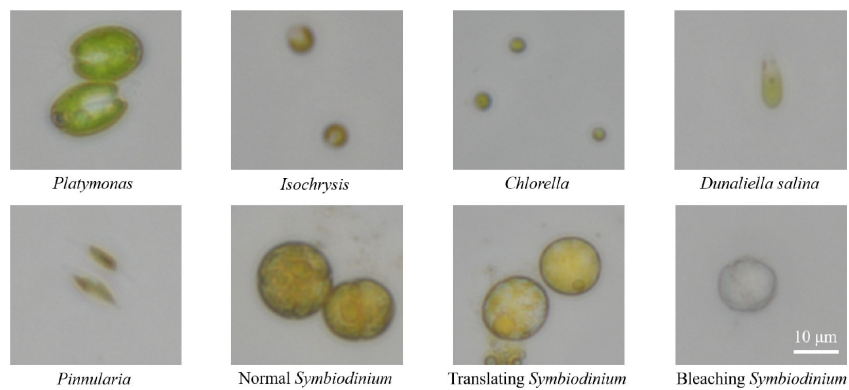

samples were taken from collected fresh seawater and then cultured in the medium.

- *Platymonas* is a genus of green algae in the family *Chlamydomonadaceae*, whose cell length is approximately 12 µm, and the algal body is flat and compressed. The cells are broadly ovate in front view, with a broad front end and a concave front of the top.
- *Isochrysis*, also known as golden algae, is a kind of algae whose active cells are 4.4–7.1 µm long and 2.7–4.4 µm wide, with a dark red, oval eye spot located in the center of the cell, occasionally near the front of the cell.
- *Chlorella* is a kind of spherical unicellular algae with a diameter of 3–8 µm, and without flagella.
- *D. salina* is a type of halophile green unicellular microalgae that is mainly found in hypersaline environments whose algal body is oval or pear-shaped, 18–28 µm long, and 9.5–14 µm wide. Without a cell wall, the front of the cell is generally concave, and there is a cup-shaped chromophore in the algae.
- *Pinnularia* is a unicellular planktonic diatom with a siliceous cell wall. The cells are spindle-shaped, with an enlarged central part and acuminate at both ends, 12–23 µm long and 2–3 µm wide. Two yellow-brown pigment bodies flank the nucleus in the center of the cell.

The algal genera mentioned above were collected from the Bohai Sea (Shangdong, China), isolated and purified by Fengyun Algae Co., Ltd., and cultured in an f/2 medium.

- *Symbiodinium*, also known as *Zooxanthella*, is a kind of algae establishing intracellular symbioses with organisms such as corals, anemones, jellyfish, nudibranchs, *Ciliophora*, *Foraminifera*, zoanthids, and sponges (Gordon and Leggat, 2010). It enters host cells by phagocytosis, persists as intracellular symbionts, multiplies, and disperses into the environment.

*Symbiodinium* used to produce this dataset was kindly donated by a scholar and isolated from the Scleractinia coral *Galaxea fascicularis* in West Island (18°14′16′′, 109°21′54′′, Sanya, Hainan, China). There were two culture mediums for *Symbiodinium*. The first culture medium was *standard* f/2 artificial seawater medium and the other in another 1-L shaker's flask was stressed by heat shock at 40°C for approximately 2 h to make the cells bleached.

Observed with a microscope, it was seen that the cells of normal *Symbiodinium* were yellowish brown, while the cells of bleached *Symbiodinium* were white, and the color of the translating cell was in between. There are minor differences in shape, size, and structure between the three groups of *Symbiodinium* cells.

## 2.2 Image acquisition and annotation

Image acquisition and processing are based on an inverted microscopic imaging platform. In our work, a high-definition camera (DS-FI3, Nikon, Japan) with a viewing field of 7.18×5.32 mm and a 5.9-megapixel complementary metal-oxide-semiconductor (CMOS) image sensor are attached to the inverted microscope. It can capture images with a resolution of up to 2,880×2,048 pixels and transfer them *via* a USB 3.0 port. The exposure time is set to 1 µs. More than 967 images of *Platymonas*, *Isochrysis*, *Chlorella*, *D. salina*, *Pinnularia*, and *Symbiodinium* have been captured individually by using a 40× microscope objective (CFI PlanFluor, NA = 0.45).

The LabelMe software, a commonly used graphical image annotation tool, was used to annotate the total dataset manually. The labeling results of different genera can be seen in Figure 5. After the annotation for the first time, another researcher proofread them using LabelMe. The original labeling results were saved as files in JSON format and were converted into txt files in YOLO annotation format for the convenience of implementing object detection algorithms. The original images were in TIFF format and were converted into PNG format.

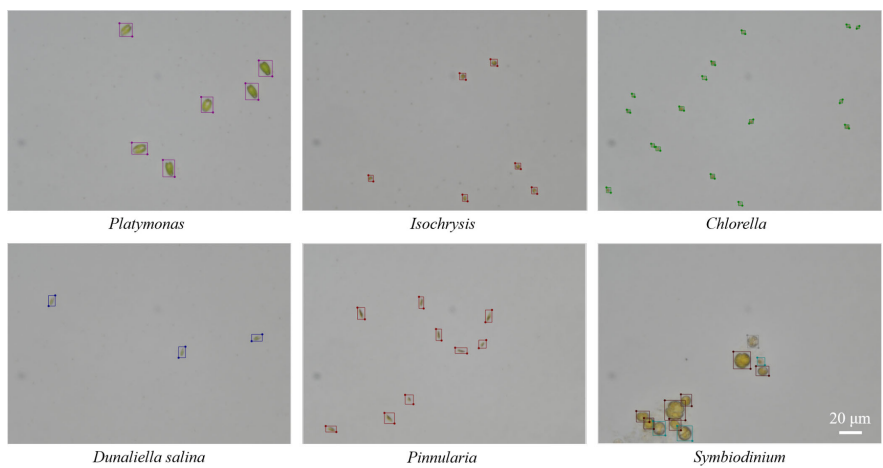

FIGURE 5
Examples of the labeling results for microalgal objects by using LabelMe software in the training set.

## 2.3 Dataset splitting

The total number of images in the dataset is 967 and all the objects in these images were annotated. The total number of annotated things is 3,915. The annotated dataset contains 150 microscopic images of each of the five genera of algae (*Platymonas*, *Isochrysis*, *Chlorella*, *D. salina*, and *Symbiodinium*), 40 images of *Pinnularia*, and 197 images of mixed samples. We divided the dataset, which included only a single class of objects per image, into a training set and a test set at a ratio of 0.7. We supplemented the test set with an image set that included multiple classes of objects per image. The total training set contained 537 images and the test set contained 430 images with an image resolution of 2,880 × 2,048, as shown in Figure 6A. The total number of microalgae objects in the training set and the test set was 2,085 and 1,830, respectively. The exact number of objects in different classes in the training and test sets is shown in Figure 6B.

Samples collected in the field often contain multiple types of microalgae; thus, we made all the mixed sample images in the test set more consistent with the actual detection conditions. In addition, the size of our test set is close to that of the training set. This is because image acquisition and labeling are time-consuming, and we hope that by training a deep learning model that performs well on a relatively small training set, we can significantly reduce the time and labor costs for researchers.

We make this dataset public with the following access address: https://github.com/Heyimace/Marine-MicroalgaeDetection-Dataset. The previous version of the dataset was used in the competition for IEEE UV2022 at https://tianchi.aliyun.com/competition/entrance/532036/introduction. The license of this dataset is GPL-3.0.

## 2.4 Implementation details

We trained and tested our dataset's classical and state-of-the-art object detection models. We hope that these experiments can also offer researchers some hints for continuing research. All experiments were done on NVIDIA GeForce RTX 3090 platforms with 24 GB RAM. The deep learning framework used is PyTorch (Paszke et al., 2017; Paszke et al., 2019). The training set in our dataset was split into two sets for training and validation. The new

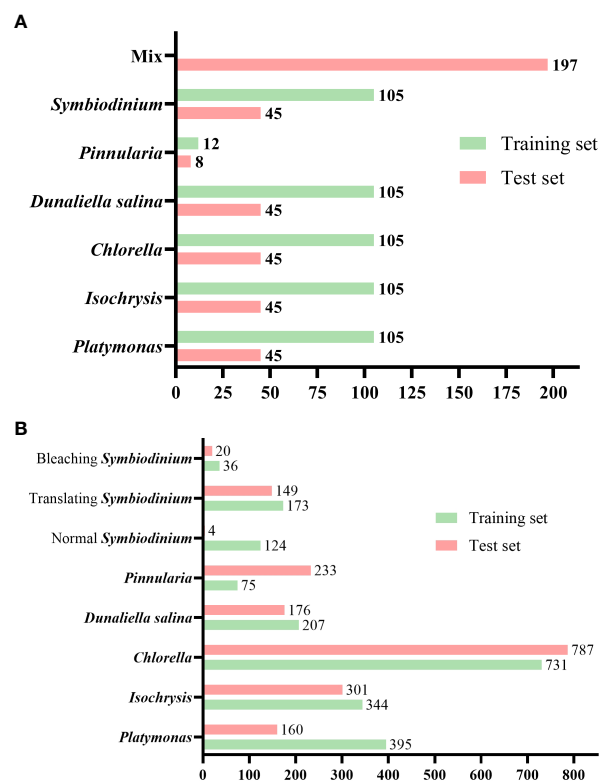

FIGURE 6
Number of images and labeled objects. (A) Number of images of various classes of algal cells in the test and training sets. (B) Number of annotated objects in the test set and training set for various classes of algal cells.

training set contains 430 images, and the validation set contains 107 images.

For Faster RCNN (Ren et al., 2017), Casacade R-CNN (Cai and Vasconcelos, 2018), Dynamic RCNN (Zhang et al., 2020b), and TOOD (Feng et al., 2021) models, we trained, validated, and tested models based on the mmdetection toolbox (Chen et al., 2019). We only used Random Flip with 0.5 probability for augmentation in the training process and Multi-Scale Flipping and Random Flipping for augmentation in the testing process. The sizes of input images in training, validation, and testing are 800 × 800. The stochastic gradient descent (SGD) optimizer was used in experiments. The learning rate was 0.01, the momentum was 0.9, and the weight decay is 0.0005. The total training epochs are 100.

For YOLOv5 and YOLOv8 models, we trained, validated, and tested models based on Ultralytics's framework (Jocher et al., 2023). The total training epochs are 100 and the training batch size is 16. The maximum detection in validation and testing is 300. The initial learning rate is 0.01, the final OneCycleLR learning rate is 0.0001, the momentum is 0.937, and mosaic augmentation is used. The fraction of image HSV-Hue augmentation is 0.015, the fraction of image HSV-Saturation augmentation is 0.7, and the fraction for image HSV-Value augmentation is 0.4. The fractions for image translation and scaling are 0.1 and 0.5, respectively. The probability of an image flipping horizontally is 0.5.

## 2.5 Evaluation metrics

Intersection over union (IoU) is a common metric in object detection. It refers to the intersection ratio between the bounding box predicted by the model and the ground truth, which is defined as:

$$IoU(A, B) = \frac{A \cap B}{A \cup B} \quad (1)$$

where $A$ and $B$ represent the area of the bounding box and ground truth predicted by the model, respectively, and the detection is considered correct when the IoU is more significant than a certain threshold.

$TP$ means True Positive prediction, $TN$ means True Negative prediction, $FP$ means False Positive prediction, and $FN$ means False Negative prediction. Then, the performance of the model can be evaluated by Precision (P), Recall (R), and Average Precision (AP), as in:

$$Precision = \frac{TP}{TP+FP} \quad (2)$$

$$Recall = \frac{TP}{TP+FN} \quad (3)$$

AP is the area under the precision–recall curve. The mean average precision (mAP) is calculated by finding the AP for each class and then the average over all classes:

$$mAP = \frac{1}{N}\sum_{i=1}^{N} AP_i \quad (4)$$

The evaluation metric in validation is mAP(0.50:0.95) to select the best model, and we gave the mAP(0.50:0.95) as well as mAP (0.50) results on the test set. The mAP(0.50:0.95) represents the average mAP over different IoU thresholds (from 0.5 to 0.95 in steps of 0.05) and the mAP(0.50) represents the average mAP over 0.5.

## 3 Results

Object detection is a computer vision task that involves locating and identifying objects in an image or video. Two main object detection methods exist: single- and two-stage object detection algorithms. One-stage object detection methods perform object classification and bounding-box regression directly without using pre-generated region proposals. Two-stage object detection methods first generate region proposals using a separate network or algorithm, such as selective search or region proposal network (RPN), and then classify and refine each proposal using another network. In this study, we trained one-stage object detection algorithms including YOLOv5, YOLOv8, and TOOD, and two-stage object detection algorithms including Faster-RCNN, Cascade-RCNN, and Dynamic-RCNN.

The detection results based on mmdetection on the test set are shown in Table 1. The evaluation metrics of the one-stage object detection algorithms are higher than the two-stage object detection algorithms. The indices of mAP (IoU = 0.5) of the Faster-RCNN series are only approximately 0.692 to 0.756. The Cascade-RCNN series, which uses cascade regression as a resampling mechanism to increase the IoU value of the proposal stage by stage so that the proposals resampled by the previous stage can adapt to the next stage with a higher threshold, have a higher detection accuracy. However, Cascade-RCNN with a larger backbone has a lower detection accuracy than the one with a smaller backbone. This may be because the larger model causes overfitting more easily. Dynamic-RCNN, which continuously adaptively increases the positive sample threshold and adaptively modifies the SmoothL1 Loss parameter, also achieves better results than Faster-RCNN. TOOD, a one-stage detection method that uses Task-aligned head and Task Alignment Learning to solve the problem of classification and positioning misalignment, achieves better results than two-stage methods.

TABLE 1   Results of different object detection methods on this dataset.

| Method | Backbone | mAP@0.5 | mAP@0.5:0.95 |
|---|---|---|---|
| Faster-RCNN (Ren et al., 2017) | ResNet50 (He et al., 2016) | 0.692 | 0.341 |
| Faster-RCNN | ResNet101 (He et al., 2016) | 0.756 | 0.407 |
| Cascade-RCNN (Cai and Vasconcelos, 2018) | ResNet50 | 0.812 | 0.484 |
| Cascade-RCNN | ResNet101 | 0.797 | 0.481 |
| Dynamic-RCNN (Zhang et al., 2020b) | ResNet50 | 0.804 | 0.478 |
| TOOD (Feng et al., 2021) | ResNet50 | 0.842 | 0.539 |
| TOOD | ResNet101 | 0.950 | 0.571 |

We also trained YOLOv5 and YOLOv8 models, two of the most advanced one-stage object detection algorithms, and explored the detection performance of models with different model sizes. The detection results of YOLOv5 and YOLOv8 on the test set are shown in Table 2, and the visualizations of detection results on the test set are shown in Figure 7. The detection results of YOLOv8m for different classes is shown in Table 3 and Figure 8. The mean mAP (IoU = 0.5:0.95) of YOLO series is higher than the other algorithms, which indicates that its prediction boxes can locate and match objects more precisely at progressively higher IoU thresholds. YOLOv8 replaces the C3 structure of YOLOv5 with the C2f structure and adjusts the number of channels for different scale models. It separates the classification and detection heads, and changes from Anchor-Based to Anchor-Free, using the TaskAlignedAssigner positive sample distribution strategy and introducing Distribution Focal Loss. Compared to YOLOv5,

YOLOv8 models lead to some improvement in our dataset. We can also find that larger models do not perform better than smaller models. In addition to the characteristics of the dataset image itself, this may also be related to the small amount of data.

## 4 Conclusion

In this paper, we apply algorithms in computer vision to the multiple object detection and physiological state assessment of marine microalgae. We introduced a new dataset for marine microalgae detection in microscopy images and the building process, including data collection, microscopic imaging, image annotation, proofreading, and splitting. This dataset contains microalgal images of six genera, namely, *Pinnularia*, *Chlorella*, *Platymonas*, *D. salina*, *Isochrysis*, and *Symbiodinium*. Moreover, we also collected the images of *Symbiodinium* in different physiological states, namely, normal, bleaching, and translating, which can be an indicator of the situation of coral and its water environment. The dataset contains 967 images, among which 537 are in the training set and 430 are in the test set.

We trained and tested a number of classical or state-of-the-art models in our dataset, including one-stage object detection algorithms and two-stage object detection algorithms. Among these algorithms, YOLOv5 and YOLOv8 were able to perform the task with high accuracy for the classification and localization of microalgal cells. The main challenges of this task can be concluded as follows:

- High-resolution images and tiny objects
- Small training set size that is close to the test set

- Multiple species in some images in the test set
- Sample imbalance in the training set

TABLE 2  Results of YOLOv5 and YOLOv8 models on this dataset.

| Method | Precision | Recall | mAP@0.5 | mAP@0.5:0.95 |
|---|---|---|---|---|
| YOLOv5s (Jocher, 2020) | 0.839 | 0.754 | 0.832 | 0.569 |
| YOLOv5m (Jocher, 2020) | 0.763 | 0.775 | 0.785 | 0.533 |
| YOLOv8s (Jocher et al., 2023) | 0.825 | 0.758 | 0.833 | 0.579 |
| YOLOv8s (Jocher et al., 2023) | 0.792 | 0.764 | 0.826 | 0.571 |

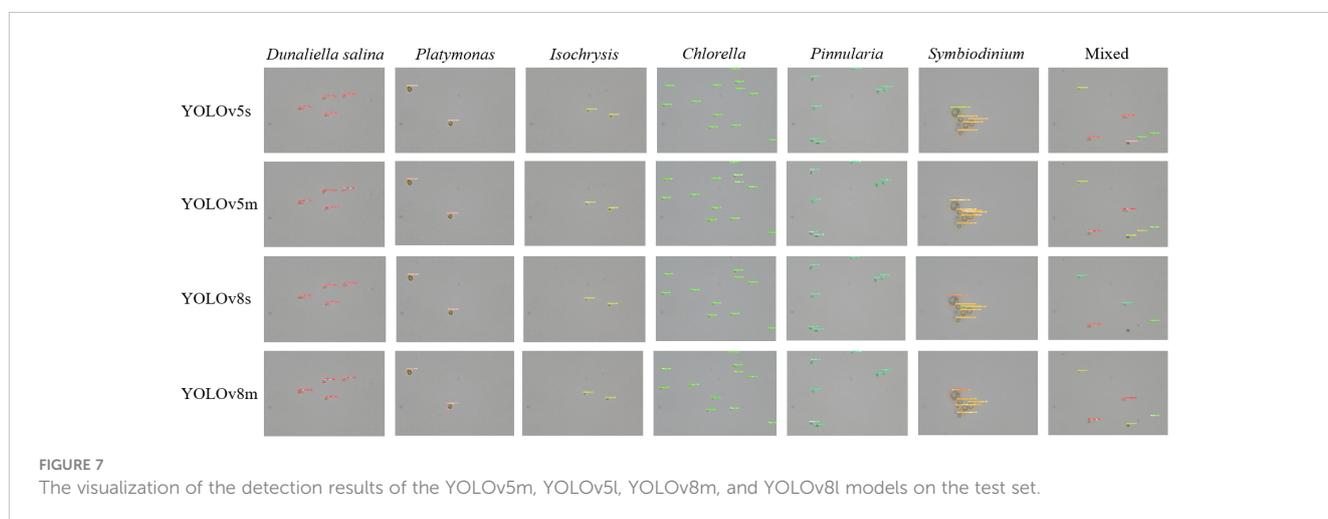

FIGURE 7
The visualization of the detection results of the YOLOv5m, YOLOv5l, YOLOv8m, and YOLOv8l models on the test set.

TABLE 3  Detection results of YOLOv8m for different classes on this dataset.

| Class | Precision | Recall | mAP@0.5 | mAP@0.5:0.95 |
|---|---|---|---|---|
| *Dunaliella salina* | 0.729 | 0.903 | 0.919 | 0.541 |
| *Platymonas* | 0.981 | 0.995 | 0.995 | 0.820 |
| *Isochrysis* | 0.766 | 0.977 | 0.907 | 0.596 |
| *Chlorella* | 0.970 | 0.863 | 0.955 | 0.609 |
| *Pinnularia* | 0.940 | 0.605 | 0.881 | 0.497 |
| Normal *Symbiodinium* | 0.388 | 0.481 | 0.505 | 0.408 |
| Translating *Symbiodinium* | 0.865 | 0.989 | 0.966 | 0.778 |
| Bleaching *Symbiodinium* | 0.697 | 0.300 | 0.483 | 0.328 |

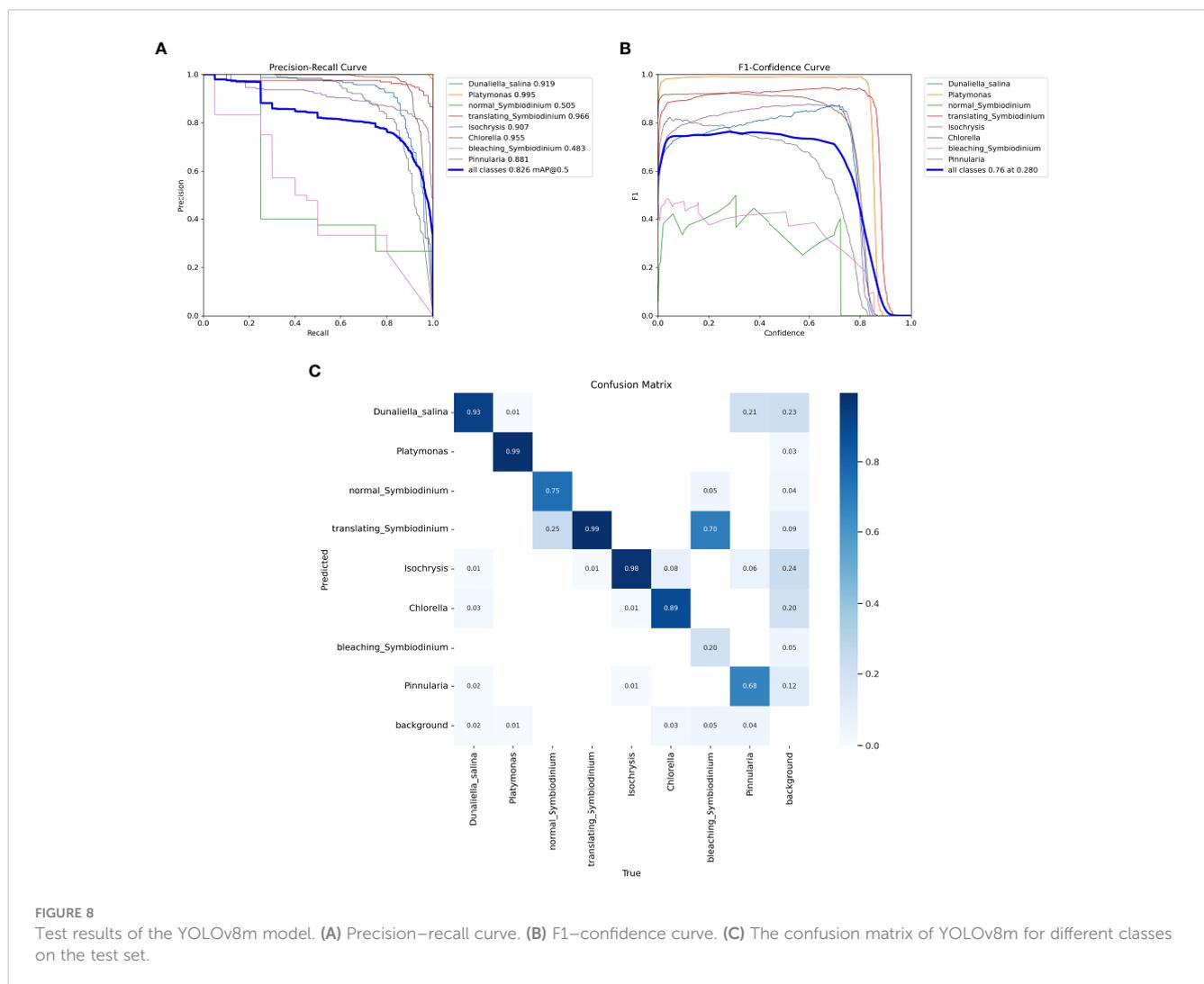

FIGURE 8
Test results of the YOLOv8m model. **(A)** Precision–recall curve. **(B)** F1–confidence curve. **(C)** The confusion matrix of YOLOv8m for different classes on the test set.

For future research, we will make a larger dataset as the next version and continue to develop stronger baselines for the microalgal detection tasks.

## Data availability statement

The datasets presented in this study can be found in online repositories. The names of the repository/repositories and accession number(s) can be found in the article/supplementary material.

## Ethics statement

Written informed consent was obtained from the individual(s) for the publication of any potentially identifiable images or data included in this article.

## Author contributions

SZ: Conceptualization, Methodology, Resources, Visualization, and Writing—Original Draft; JJ: Investigation, Software, and Writing—Original Draft; XH: Data Curation; PF: Project Administration and Funding Acquisition; HY: Writing—Review and Editing, Supervision, and Funding Acquisition. All authors contributed to the article and approved the submitted version.

## Funding

This work is supported by the Hainan Provincial Key Research and Development Program (ZDYF2020026), the Hainan Provincial Higher Education Scientific Research Project Key Project (HNKY2021ZD-7), the Hainan Provincial Social Science Research Project (HNSK(YB)-20-24), and the Hainan Provincial Natural Science Foundation (620MS022).

## Conflict of interest

The authors declare that the research was conducted in the absence of any commercial or financial relationships that could be construed as a potential conflict of interest.

## Publisher's note

All claims expressed in this article are solely those of the authors and do not necessarily represent those of their affiliated organizations, or those of the publisher, the editors and the reviewers. Any product that may be evaluated in this article, or claim that may be made by its manufacturer, is not guaranteed or endorsed by the publisher.